# MMRAG: Multi-Mode Retrieval-Augmented Generation with Large Language Models for Biomedical In-Context Learning

Zaifu Zhan, MEng[1], Jun Wang, PhD[2], Shuang Zhou, PhD[2], Jiawen Deng[3], Rui Zhang, PhD[2*]

[1]Department of Electrical and Computer Engineering, University of Minnesota, Minneapolis, MN, USA

[2]Division of Computational Health Sciences, Department of Surgery, University of Minnesota, Minneapolis, Minnesota, USA,

[3]Department of Computer Science and Engineering, University of Minnesota, Minneapolis, MN, USA

## ABSTRACT

**Objective**: To optimize in-context learning in biomedical natural language processing by improving example selection.

**Methods**: We introduce a novel multi-mode retrieval-augmented generation (MMRAG) framework, which integrates four retrieval strategies: (1) Random Mode, selecting examples arbitrarily; (2) Top Mode, retrieving the most relevant examples based on similarity; (3) Diversity Mode, ensuring variation in selected examples; and (4) Class Mode, selecting category-representative examples. This study evaluates MMRAG on three core biomedical NLP tasks: Named Entity Recognition (NER), Relation Extraction (RE), and Text Classification (TC). The datasets used include BC2GM for gene and protein mention recognition (NER), DDI for drug-drug interaction extraction (RE), GIT for general biomedical information extraction (RE), and HealthAdvice for health-related text classification (TC). The framework is tested with two large language models (Llama-2-7B, Llama-3-8B) and three retrievers (Contriever, MedCPT, BGE-Large) to assess performance across different retrieval strategies.

**Results**: The results from the Random mode indicate that providing more examples in the prompt improves the model's generation performance. Meanwhile, Top mode and Diversity mode significantly outperform Random mode on the RE (DDI) task, achieving an F1 score of 0.9669—a 26.4% improvement. Among the three retrievers tested, Contriever outperformed the other two in a greater number of experiments. Additionally, Llama 2 and Llama 3 demonstrated varying capabilities across different tasks, with Llama 3 showing a clear advantage in handling NER tasks.

**Conclusion**: MMRAG effectively enhances biomedical in-context learning by refining example selection, mitigating data scarcity issues, and demonstrating superior adaptability for NLP-driven healthcare applications.

# 1 Introduction

Large language models (LLMs) such as the GPT series[1,2] and Llama models[3], are now playing an increasingly pivotal role in healthcare[4], demonstrating remarkable capabilities in processing biomedical text and electronic health records (EHRs)[5–7]. Their applications include extracting clinical insights[8,9], supporting patient monitoring[10], performing clinical text classification[11], addressing natural language inference[12], and facilitating medical information extraction[13–15].

However, due to the need to protect patient privacy[16], research in medical NLP faces significant challenges related to insufficient training data[17,18]. Although hospitals and healthcare institutions often possess large amounts of raw EHR data, much of this data is unstructured and unlabeled[4]. The process of curating, annotating, and structuring this data into usable formats for training machine learning models is time-consuming, money-consuming and resource-intensive. Furthermore, for rare diseases[19,20]—such as genetic disorders, orphan diseases, and conditions with limited geographic or demographic prevalence—the lack of documented cases exacerbates the data scarcity issue. These conditions inherently have fewer reported instances, making it difficult to collect even raw EHR data, let alone labeled datasets. Finally, privacy regulations such as HIPAA or GDPR further restrict access to sensitive patient information, making it even harder to obtain and share data for research purposes. As a result, novel approaches are urgently needed to address these challenges.

In this context, in-context learning (ICL) might be a promising solution. Along with the development of LLMs, the size of these models has grown significantly, the amount of training data has increased, and the length of prompts has expanded[21]. As a result, LLMs have begun to demonstrate the ability to learn from context, a capability known as ICL[22]. This approach allows the model to learn patterns from a few examples provided within the prompt, enabling it to deliver higher-quality responses. In-context learning has several advantages. First, it eliminates the need for training or fine-tuning, which requires substantial computational resources and large amounts of data; instead, ICL only requires a few examples[23]. Second, designing prompts is much simpler than training a model[24], making this approach accessible to individuals without specialized knowledge of LLMs, such as doctors in hospitals. Third, providing examples aligns with the human brain's natural way of thinking—learning through analogy. This paradigm allows for the seamless incorporation of human knowledge into LLMs by simply adjusting demonstrations and templates[25,26]. Finally, a single model can address a wide range of tasks by designing different prompts and providing corresponding examples, showcasing its remarkable versatility[27]. Therefore, ICL is indispensable in the field of healthcare.

By leveraging the ability of large language models to learn patterns from a few carefully designed examples embedded in the prompt, ICL could mitigate the insufficient data challenges in the biomedical domain. However, despite its potential, the process of selecting relevant examples from existing verified or labeled data for ICL remains underexplored[28]. The careful curation of examples is crucial in biomedical contexts, as the examples not only encapsulate rich medical knowledge—such as clinical guidelines, rare disease manifestations, or treatment pathways—but also significantly influence the quality and accuracy of the model's responses based on their relevance to the input[23,25,28]. Therefore, developing effective methodologies for selecting high-quality examples becomes essential to unlocking the full potential of ICL in medical NLP.

Fortunately, Retrieval-Augmented Generation (RAG)[29,30], a technique designed to extract information directly relevant to an input query, offers a promising approach to address this challenge. RAG can enhance the example selection process by efficiently retrieving contextually similar data points[31,32], making it a natural complement to ICL. To further advance the application of ICL in medical NLP, we propose an innovative MMRAG framework that leverages RAG to extract and select similar examples using different strategies tailored to diverse use cases. This framework introduces four distinct modes for example selection: random mode, where examples are selected without any specific prioritization; top mode, which retrieves the most relevant examples based on similarity metrics; diverse mode, which ensures a broad range of examples to increase diversity; and classes mode, which focuses on selecting examples representative of specific categories or conditions. Each mode corresponds to a unique rule for selecting examples, providing flexibility and adaptability to a variety of medical NLP tasks. By integrating these strategies, the proposed framework offers a systematic and effective approach to maximize the potential of ICL in addressing critical challenges in medical NLP, particularly in domains constrained by data scarcity and privacy considerations. Our key contributions are summarized as follows:

- To the best of our knowledge, this is the first study to investigate the selection of examples in the biomedical domain for in-context learning.

- We propose a novel framework with four distinct modes (random, top, diverse, classes) to optimize example selection and explore the potential of ICL in biomedical contexts.

- We conduct extensive experiments, evaluating the framework across two LLMs on four datasets with three retrievers, providing comprehensive insights into its performance.

## 2 Methods

### 2.1 Overview of methods

This study introduces MMRAG (Multi-mode Retrieval-Augmented Generation) framework, designed to enhance ICL in biomedical NLP by optimizing example selection. MMRAG employs four retrieval strategies: Random Mode (arbitrary selection), Top Mode (most similar examples), Diversity Mode (balanced similarity and variation), and Class Mode (representative examples from different categories). The framework integrates RAG with LLMs (Llama-2-7B[33], Llama-3-8B[34]) and three retrievers (Contriever[35], MedCPT[36], BGE-Large[37]) to systematically evaluate retrieval effectiveness. Our experiments cover Named Entity Recognition (NER), Relation Extraction (RE), and Text Classification (TC) using four biomedical datasets (BC2GM[38], DDI[39], GIT[40], HealthAdvice[41]). This framework demonstrates how tailored retrieval strategies can significantly improve ICL performance in biomedical NLP, addressing challenges such as data scarcity and privacy constraints.

### 2.2 Tasks and datasets

In this work, we selected three core information extraction tasks in biomedical domains: NER, RE, TC. The main purpose of these tasks is to extract the useful information from the raw text, which is process from unstructured data to useful information for any downstream applications.

**Named Entity Recognition (NER) - BC2GM**[38]: The BC2GM dataset is a widely used benchmark for biomedical named entity recognition (NER), specifically focusing on gene and protein mentions in scientific literature. Originating from the BioCreative II Gene Mention task, it consists of manually annotated gene mentions extracted from PubMed abstracts. This dataset plays a pivotal role in biomedical text mining by enabling the development and evaluation of NER models that can accurately identify gene-related entities.

**Relation Extraction (RE) - DDI**[39]: The DDI (Drug-Drug Interaction) dataset is a key resource for relation extraction in the biomedical domain, particularly designed to identify and classify drug-drug interactions from text. It includes sentences from biomedical literature and drug product information sources, annotated with drug entities and their interactions. This dataset is crucial for pharmacovigilance and drug safety, as accurate extraction of drug interactions supports clinical decision-making, prevents adverse drug reactions, and enhances patient safety.

**Relation Extraction (RE) - GIT**[40]: The GIT (General BioMedical and Complementary and Integrative Health Triples) dataset is a high-quality biomedical triple extraction dataset specifically focused on non-drug therapies. It is characterized by its high-quality annotations and comprehensive coverage of relation types, making it a valuable resource for biomedical relation extraction. The dataset includes 22 relation types derived from SemMedDB, providing structured representations of relationships within biomedical texts. By supporting the extraction of meaningful entity-relation triples, GIT facilitates advancements in knowledge graph construction, automated reasoning, and biomedical text mining applications related to non-drug interventions.

**Text Classification (TC) - HealthAdvice**[41]: The HealthAdvice dataset is designed for text classification tasks related to health information and advisory content. It consists of a diverse collection of health-related advice. The dataset is structured to facilitate automatic classification of health advice into relevant categories, supporting applications such as misinformation detection, personalized health recommendations, and automated triaging of medical inquiries. By leveraging this dataset, researchers can enhance natural language processing (NLP) models tailored for health communication and decision support.

The combination of these datasets—BC2GM for entity recognition, DDI and GIT for relation extraction, and HealthAdvice for text classification—provides a comprehensive foundation for biomedical NLP research. These datasets enable the development of advanced NLP models capable of extracting critical biomedical knowledge from vast textual sources. By improving the accuracy of entity recognition, relationship extraction, and health-related text classification, they contribute significantly to biomedical informatics, clinical decision support, and evidence-based medicine.

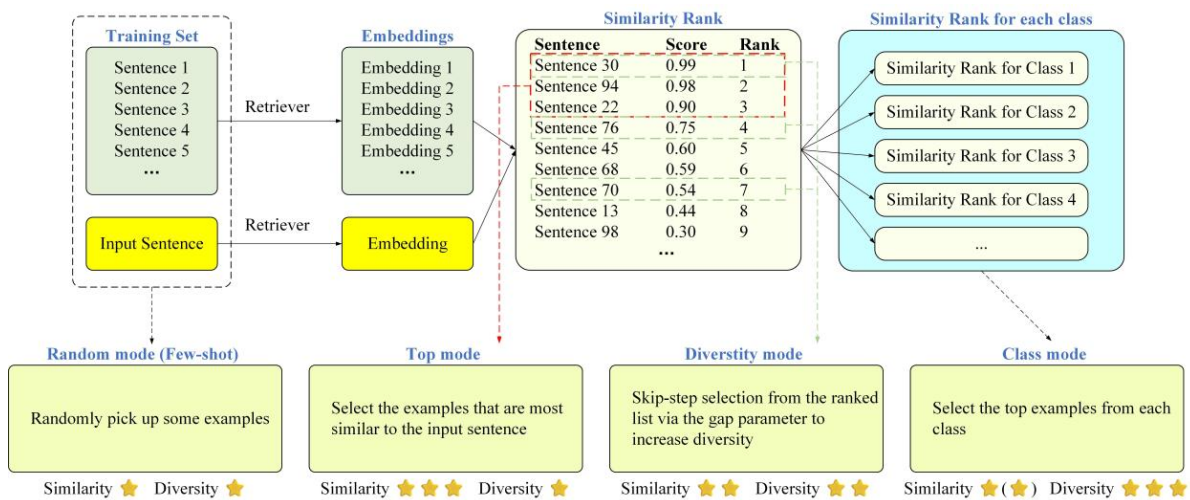

Figure 1. An overview of the multi-mode retrieval-augmented generation framework.

## 2.2 MMRAG Framework

The motivation for the proposed framework is to release the potential of ICL using the RAG technique for biomedical information extraction with LLMs. As depicted in Fig. 1, the retrieval process begins by embedding both the training set sentences and the input sentence. A retriever then ranks the training sentences based on similarity scores. The selection mode determines how the final examples are chosen from this ranked list. Then, we introduced four different modes for selecting examples: Random Mode, Top Mode, Diversity Mode, and Class Mode. These modes determine how retrieved sentences are selected based on their similarity rankings, striking different balances between similarity and diversity.

**Random Mode (Few-shot):** Random Mode selects examples arbitrarily from the training set without considering similarity ranking. This introduces randomness, which can improve model generalization but may result in irrelevant examples. While this mode maximizes diversity, it does not ensure the selection of highly relevant sentences.

**Top Mode:** Top Mode selects the most similar examples by directly choosing the top-ranked sentences. This ensures high contextual similarity between the input sentence and the retrieved examples. However, it may lead to redundancy, as the examples are often very similar to each other, limiting the diversity of information.

**Diversity Mode:** Diversity Mode introduces a skip-step selection mechanism, where examples are picked at intervals from the ranked list instead of consecutively selecting the top sentences. By skipping certain highly ranked examples, this method ensures that the retrieved examples maintain relevance while increasing the diversity of information.

**Class Mode:** Class Mode selects the most relevant example from each predefined semantic class rather than purely ranking by similarity. This ensures that different types of sentences are represented in the final selection, improving coverage across multiple categories. While it maintains relevance, some selected examples may have slightly lower similarity scores compared to those chosen in Top Mode.

Each mode provides a different trade-off between similarity and diversity. Random Mode cannot guarantee diversity and relevance, but it represents the most common method. Top Mode guarantees high similarity but lacks diversity. Diversity Mode balances both by selecting non-consecutive similar examples. Class Mode further enhances diversity by ensuring category coverage. Choosing the appropriate mode depends on the specific requirements of the retrieval task.

## 2.3 Experimental Settings

Our experiments were conducted using two large language models, Llama-2-7B[33] and Llama-3-8B[34], both of which have demonstrated strong performance in natural language processing tasks. To retrieve relevant examples, we employed three different retrievers: Contriever[35], a dense retriever trained with contrastive learning for general-purpose retrieval; MedCPT[36], a retriever optimized for medical and clinical text retrieval; and BGE-Large[37], an embedding-based retriever known for its effectiveness in similarity search. These retrievers were chosen to test the adaptability of our framework across different retrieval strategies.

All experiments were performed on NVIDIA A100 GPUs with 40GB memory, ensuring sufficient computational resources for training and inference. The batch size per device was set to 4 for both training and evaluation, and the inference was carried out sentence by sentence to optimize efficiency. For automated evaluation, we applied Low-Rank Adaptation[42], a parameter-efficient fine-tuning method, with a rank of 64, alpha set to 32, and a dropout rate of 0.1, to fine-tune the model. The models were optimized using the AdamW optimizer with a learning rate of 1e-5. Fine-tuning was conducted for 5,000 steps, with evaluations performed every 1,000 steps, and the best-performing model was selected for inference.

For evaluation, we adopted Micro Precision, Recall, and F1-score, in line with established studies. A prediction was considered correct only if the entire output exactly matched the ground truth, ensuring a strict and reliable assessment of model performance. This evaluation approach allowed us to measure the effectiveness of different retrieval strategies and selection modes in a controlled and reproducible manner.

## 3 Results

### 3.1 Performance in Random Mode

In Random Mode, examples were selected arbitrarily from the dataset, providing a baseline comparison for retrieval-enhanced selection strategies. Table 1 shows the mean and standard deviation of F1 scores for different tasks and datasets since the random mode introduces huge randomness. For the NER task on BC2GM, Llama-2-7B achieved an F1 score of 0.8766 with one example, improving to 0.9172 with ten examples. Llama-3-8B showed superior performance, reaching 0.9660 with one example and further increasing to 0.9782 with ten examples. In the RE task on the DDI dataset, Llama-2-7B fluctuated between 0.7049 and 0.7050 across different example counts, while Llama-3-8B exhibited marginal improvement, achieving a maximum F1 score of 0.7114 with five examples. In text classification on HealthAdvice, Llama-2-7B increased from 0.8954 (one example) to 0.9171 (ten examples), while Llama-3-8B ranged from 0.8848 to 0.9084. These results indicate that while Random Mode can yield reasonable performance improvements, it lacks the reliability and control of retrieval-based selection.

**Table 1.** Comparison of two LLMs across three datasets in Random Mode. The reported values represent the mean and standard deviation over three repeated experiments

| Tasks | Datasets | Models | Examples | Precision | Recall | F1 |
|-------|----------|--------|----------|-----------|--------|-----|
| NER | BC2GM | Llama-2-7B | 1 | $0.9232 \pm 0.0001$ | $0.8345 \pm 0.0007$ | $0.8766 \pm 0.0004$ |
| | | | 5 | $0.9267 \pm 0.0008$ | $0.8878 \pm 0.0032$ | $0.9069 \pm 0.0018$ |
| | | | 10 | $0.9318 \pm 0.0007$ | $0.9032 \pm 0.0006$ | $0.9172 \pm 0.0006$ |
| | | Llama-3-8B | 1 | $0.9733 \pm 0.0006$ | $0.9589 \pm 0.0008$ | $0.9660 \pm 0.0001$ |
| | | | 5 | $0.9806 \pm 0.0001$ | $0.9712 \pm 0.0004$ | $0.9759 \pm 0.0003$ |
| | | | 10 | $0.9814 \pm 0.0007$ | $0.9749 \pm 0.0010$ | $0.9782 \pm 0.0004$ |

| Tasks | Datasets | | Num | | | |
|---|---|---|---|---|---|---|
| RE | DDI | Llama-2-7B | 1 | 0.7049 ± 0.0031 | 0.7049 ± 0.0031 | 0.7049 ± 0.0031 |
| | | | 5 | 0.7081 ± 0.0002 | 0.7081 ± 0.0002 | 0.7081 ± 0.0002 |
| | | | 10 | 0.7036 ± 0.0128 | 0.7064 ± 0.0083 | 0.7050 ± 0.0105 |
| | | Llama-3-8B | 1 | 0.7004 ± 0.0018 | 0.7004 ± 0.0018 | 0.7004 ± 0.0018 |
| | | | 5 | 0.7114 ± 0.0045 | 0.7114 ± 0.0045 | 0.7114 ± 0.0045 |
| | | | 10 | 0.7005 ± 0.0066 | 0.7005 ± 0.0066 | 0.7005 ± 0.0066 |
| TC | HealthAdvice | Llama-2-7B | 1 | 0.8954 ± 0.0027 | 0.8954 ± 0.0027 | 0.8954 ± 0.0027 |
| | | | 5 | 0.9154 ± 0.0017 | 0.9159 ± 0.0020 | 0.9156 ± 0.0018 |
| | | | 10 | 0.9171 ± 0.0021 | 0.9171 ± 0.0021 | 0.9171 ± 0.0021 |
| | | Llama-3-8B | 1 | 0.8848 ± 0.0025 | 0.8848 ± 0.0025 | 0.8848 ± 0.0025 |
| | | | 5 | 0.9040 ± 0.0019 | 0.9040 ± 0.0019 | 0.9040 ± 0.0019 |
| | | | 10 | 0.9084 ± 0.0006 | 0.9084 ± 0.0006 | 0.9084 ± 0.0006 |

## 3.2 Performance in Top Mode

Top Mode retrieves the most similar examples based on similarity rankings. Table 2 presents the results for this mode, demonstrating notable performance improvements over Random Mode. In NER, Llama-2-7B reached an F1 score of 0.9098 with five examples when using the Contriever retriever, while Llama-3-8B achieved a peak score of 0.9726. The MedCPT retriever also contributed to strong results, with Llama-3-8B scoring 0.9706. For RE on the DDI dataset, Contriever retrieval led to the highest F1 score of 0.9669 for Llama-2-7B and 0.9573 for Llama-3-8B. Similarly, for text classification, Top Mode improved performance, with Llama-2-7B reaching 0.8969 and Llama-3-8B achieving 0.8923 when using Contriever. These results confirm that retrieval-driven selection significantly enhances model effectiveness, particularly when using domain-optimized retrievers.

**Table 2.** Comparison of two LLMs across three datasets using three retrievers in Top Mode.

| Tasks | Datasets | Retrievers | Num | Llama 2 | | | Llama 3 | | |
|---|---|---|---|---|---|---|---|---|---|
| | | | | Precision | Recall | F1 | Precision | Recall | F1 |

| Task | Dataset | Retriever | k | | | | | | |
|------|---------|-----------|---|---|---|---|---|---|---|
| NER | BC2GM | BGE-Large | 1 | 0.9187 | 0.7310 | 0.8142 | 0.9573 | 0.9365 | 0.9468 |
| | | | 5 | 0.9154 | 0.7318 | 0.8134 | 0.9080 | 0.8820 | 0.8948 |
| | | | 10 | 0.9130 | 0.7197 | 0.8049 | 0.9052 | 0.8763 | 0.8905 |
| | | Contriever | 1 | 0.9030 | 0.7755 | 0.8344 | 0.9468 | 0.9180 | 0.9322 |
| | | | 5 | 0.9242 | 0.8958 | 0.9098 | 0.9747 | 0.9686 | 0.9716 |
| | | | 10 | 0.9167 | 0.8633 | 0.8892 | 0.9780 | 0.9672 | 0.9726 |
| | | MedCPT | 1 | 0.8608 | 0.7302 | 0.7902 | 0.9500 | 0.9204 | 0.9350 |
| | | | 5 | 0.9324 | 0.8527 | 0.8908 | 0.9784 | 0.9629 | 0.9706 |
| | | | 10 | 0.9228 | 0.8828 | 0.9024 | 0.9803 | 0.9493 | 0.9646 |
| RE | DDI | BGE-Large | 1 | 0.7358 | 0.7358 | 0.7358 | 0.7305 | 0.7305 | 0.7305 |
| | | | 5 | 0.7745 | 0.7745 | 0.7745 | 0.7589 | 0.7589 | 0.7589 |
| | | | 10 | 0.8026 | 0.8026 | 0.8026 | 0.7795 | 0.7795 | 0.7795 |
| | | Contriever | 1 | 0.9662 | 0.9662 | 0.9662 | 0.7532 | 0.7550 | 0.7541 |
| | | | 5 | 0.9504 | 0.9649 | 0.9576 | 0.9646 | 0.9646 | 0.9646 |
| | | | 10 | 0.9669 | 0.9669 | 0.9669 | 0.9573 | 0.9573 | 0.9573 |
| | | MedCPT | 1 | 0.8146 | 0.8149 | 0.8148 | 0.7050 | 0.7106 | 0.7078 |
| | | | 5 | 0.8152 | 0.8152 | 0.8152 | 0.8205 | 0.8205 | 0.8205 |
| | | | 10 | 0.8264 | 0.8264 | 0.8264 | 0.8175 | 0.8175 | 0.8175 |
| TC | HealthAdvice | BGE-Large | 1 | 0.8836 | 0.8836 | 0.8836 | 0.8180 | 0.8180 | 0.8180 |
| | | | 5 | 0.8906 | 0.8906 | 0.8906 | 0.8612 | 0.8612 | 0.8612 |
| | | | 10 | 0.8998 | 0.8998 | 0.8998 | 0.8646 | 0.8646 | 0.8646 |
| | | Contriever | 1 | 0.8900 | 0.8900 | 0.8900 | 0.8116 | 0.8116 | 0.8116 |
| | | | 5 | 0.8892 | 0.8969 | 0.8930 | 0.8923 | 0.8923 | 0.8923 |
| | | | 10 | 0.8969 | 0.8969 | 0.8969 | 0.8571 | 0.8606 | 0.8589 |
| | | MedCPT | 1 | 0.7454 | 0.7454 | 0.7454 | 0.7028 | 0.7028 | 0.7028 |
| | | | 5 | 0.7999 | 0.8128 | 0.8063 | 0.8243 | 0.8243 | 0.8243 |
| | | | 10 | 0.7797 | 0.7869 | 0.7833 | 0.7954 | 0.8105 | 0.8029 |

## 3.3 Performance in Diversity Mode

Diversity Mode introduces a balance between similarity and variation by selecting examples at intervals within the ranked retrieval list. Table 3 presents the results, demonstrating that this approach improved model robustness across datasets. In NER, the highest F1 scores were achieved using the Contriever retriever, with Llama-2-7B reaching 0.9134 and Llama-3-8B achieving 0.9731 with ten examples. For the DDI dataset, Diversity Mode yielded stable results, with Llama-2-7B reaching 0.9666 and Llama-3-8B scoring 0.9623 using Contriever retrieval. Text classification results remained competitive, with Llama-2-7B achieving an F1 score of 0.8980 and Llama-3-8B reaching 0.8765. The results indicate that Diversity Mode helps mitigate overfitting while maintaining high task relevance.

**Table 3.** Comparison of two LLMs across three datasets using three retrievers in Diversity Mode.

| Tasks and datasets | Retrievers | Gap | Llama2 (5 examples) | | | Llama2 (10 examples) | | | Llama3 (5 examples) | | | Llama3 (10 examples) | | |
|---|---|---|---|---|---|---|---|---|---|---|---|---|---|---|
| | | | Precision | Recall | F1 | Precision | Recall | F1 | Precision | Recall | F1 | Precision | Recall | F1 |
| NER, BC2GM | BGE-Large | 1 | 0.9154 | 0.7318 | 0.8134 | 0.9130 | 0.7197 | 0.8049 | 0.9080 | 0.8820 | 0.8948 | 0.9052 | 0.8763 | 0.8905 |
| | | 2 | 0.9145 | 0.7360 | 0.8156 | 0.9147 | 0.7235 | 0.8079 | 0.9109 | 0.8815 | 0.8960 | 0.9059 | 0.8762 | 0.8908 |
| | | 3 | 0.9125 | 0.7415 | 0.8182 | 0.9133 | 0.7192 | 0.8047 | 0.9081 | 0.8828 | 0.8953 | 0.9052 | 0.8797 | 0.8923 |
| | Contriever | 1 | 0.9242 | 0.8958 | 0.9098 | 0.9167 | 0.8633 | 0.8892 | 0.9747 | 0.9686 | 0.9716 | 0.9780 | 0.9672 | 0.9726 |
| | | 2 | 0.9292 | 0.8980 | 0.9134 | 0.9192 | 0.8710 | 0.8945 | 0.9712 | 0.9710 | 0.9711 | 0.9788 | 0.9675 | 0.9731 |
| | | 3 | 0.9288 | 0.8798 | 0.9036 | 0.9168 | 0.8921 | 0.9043 | 0.9747 | 0.9636 | 0.9691 | 0.9782 | 0.9688 | 0.9735 |
| | MedCPT | 1 | 0.9324 | 0.8527 | 0.8908 | 0.9228 | 0.8828 | 0.9024 | 0.9784 | 0.9629 | 0.9706 | 0.9803 | 0.9493 | 0.9646 |
| | | 2 | 0.9411 | 0.8386 | 0.8869 | 0.9206 | 0.8615 | 0.8900 | 0.9780 | 0.9601 | 0.9690 | 0.9811 | 0.9520 | 0.9663 |
| | | 3 | 0.9408 | 0.8272 | 0.8804 | 0.9159 | 0.8481 | 0.8807 | 0.9751 | 0.9440 | 0.9593 | 0.9762 | 0.9634 | 0.9698 |
| RE, DDI | BGE-Large | 1 | 0.7745 | 0.7745 | 0.7745 | 0.8026 | 0.8026 | 0.8026 | 0.7589 | 0.7589 | 0.7589 | 0.7795 | 0.7795 | 0.7795 |
| | | 2 | 0.7811 | 0.7811 | 0.7811 | 0.7768 | 0.7768 | 0.7768 | 0.7450 | 0.7450 | 0.7450 | 0.7385 | 0.7385 | 0.7385 |
| | | 3 | 0.7666 | 0.7666 | 0.7666 | 0.7611 | 0.7611 | 0.7611 | 0.7440 | 0.7440 | 0.7440 | 0.7487 | 0.7487 | 0.7487 |
| | Contriever | 1 | 0.9504 | 0.9649 | 0.9576 | 0.9669 | 0.9669 | 0.9669 | 0.9646 | 0.9646 | 0.9646 | 0.9573 | 0.9573 | 0.9573 |
| | | 2 | 0.9666 | 0.9666 | 0.9666 | 0.9669 | 0.9669 | 0.9669 | 0.9623 | 0.9623 | 0.9623 | 0.9334 | 0.9334 | 0.9334 |
| | | 3 | 0.9659 | 0.9659 | 0.9659 | 0.9649 | 0.9662 | 0.9656 | 0.9613 | 0.9613 | 0.9613 | 0.8574 | 0.8583 | 0.8579 |
| | MedCPT | 1 | 0.8152 | 0.8152 | 0.8152 | 0.8264 | 0.8264 | 0.8264 | 0.8205 | 0.8205 | 0.8205 | 0.8175 | 0.8175 | 0.8175 |
| | | 2 | 0.8199 | 0.8199 | 0.8199 | 0.8170 | 0.8170 | 0.8170 | 0.8136 | 0.8136 | 0.8136 | 0.8228 | 0.8228 | 0.8228 |
| | | 3 | 0.8252 | 0.8252 | 0.8252 | 0.8211 | 0.8211 | 0.8211 | 0.8162 | 0.8162 | 0.8162 | 0.8319 | 0.8319 | 0.8319 |
| TC, HealthAdvice | BGE-Large | 1 | 0.8906 | 0.8906 | 0.8906 | 0.8998 | 0.8998 | 0.8998 | 0.8612 | 0.8612 | 0.8612 | 0.8646 | 0.8646 | 0.8646 |
| | | 2 | 0.8796 | 0.8796 | 0.8796 | 0.8940 | 0.8940 | 0.8940 | 0.8606 | 0.8606 | 0.8606 | 0.8445 | 0.8445 | 0.8445 |
| | | 3 | 0.8894 | 0.8894 | 0.8894 | 0.8980 | 0.8980 | 0.8980 | 0.8583 | 0.8583 | 0.8583 | 0.8623 | 0.8623 | 0.8623 |
| | Contriever | 3 | 0.8888 | 0.8934 | 0.8911 | 0.8899 | 0.8899 | 0.8899 | 0.9062 | 0.9067 | 0.9064 | 0.8762 | 0.8767 | 0.8765 |
| | | 1 | 0.8892 | 0.8969 | 0.8930 | 0.8969 | 0.8969 | 0.8969 | 0.8923 | 0.8923 | 0.8923 | 0.8571 | 0.8606 | 0.8589 |
| | | 2 | 0.8779 | 0.8825 | 0.8802 | 0.8928 | 0.8928 | 0.8928 | 0.9026 | 0.9026 | 0.9026 | 0.8562 | 0.8675 | 0.8618 |
| | MedCPT | 1 | 0.7999 | 0.8128 | 0.8063 | 0.7797 | 0.7869 | 0.7833 | 0.8243 | 0.8243 | 0.8243 | 0.7954 | 0.8105 | 0.8029 |
| | | 2 | 0.8179 | 0.8203 | 0.8191 | 0.8154 | 0.8173 | 0.8164 | 0.8324 | 0.8324 | 0.8324 | 0.8397 | 0.8416 | 0.8406 |
| | | 3 | 0.8084 | 0.8116 | 0.8100 | 0.7522 | 0.7752 | 0.7636 | 0.8101 | 0.8111 | 0.8106 | 0.8393 | 0.8393 | 0.8393 |

## 3.4 Performance in Class Mode

Table 4 compares Llama2 and Llama3 across three datasets using three retrievers in Class Mode. Llama3 generally outperforms Llama2, especially in RE (GIT) where its F1-score with MedCPT reaches 0.6129, significantly higher than Llama2's 0.3786. In RE (DDI), Contriever achieves the best performance for both models, with Llama3 scoring 0.7060. For TC (HealthAdvice) both models perform well, with Llama2 slightly ahead using random retrieval, but Llama3 excels with Contriever (0.8917 F1-score). These results highlight Llama3's superior retrieval utilization and the impact of effective retrievers like MedCPT and Contriever on model performance.

**Table 4.** Comparison of two LLMs across three datasets using three retrievers in Class Mode.

| Tasks, Datasets, Examples | Retrievers | Llama 2 | Llama 3 |
|---|---|---|---|

|  |  | Precision | Recall | F1 | Precision | Recall | F1 |
|---|---|---|---|---|---|---|---|
| RE, DDI, examples:4 | Random | 0.6725 | 0.6725 | 0.6725 | 0.6738 | 0.6738 | 0.6738 |
|  | MedCPT | 0.6825 | 0.6825 | 0.6825 | 0.6841 | 0.6841 | 0.6841 |
|  | Contriever | 0.6915 | 0.6917 | 0.6916 | 0.7060 | 0.7060 | 0.7060 |
|  | BGE-Large | 0.6739 | 0.6755 | 0.6747 | 0.6881 | 0.6881 | 0.6881 |
| RE, GIT, examples:22 | Random | 0.3004 | 0.3004 | 0.3004 | 0.5075 | 0.5075 | 0.5075 |
|  | MedCPT | 0.3786 | 0.3786 | 0.3786 | 0.6129 | 0.6129 | 0.6129 |
|  | Contriever | 0.2354 | 0.2354 | 0.2354 | 0.5978 | 0.5978 | 0.5978 |
|  | BGE-Large | 0.3143 | 0.3143 | 0.3143 | 0.6108 | 0.6108 | 0.6108 |
| TC, HealthAdvice, examples:3 | Random | 0.8733 | 0.8733 | 0.8733 | 0.8658 | 0.8658 | 0.8658 |
|  | MedCPT | 0.8986 | 0.8986 | 0.8986 | 0.8842 | 0.8842 | 0.8842 |
|  | Contriever | 0.8940 | 0.8940 | 0.8940 | 0.8917 | 0.8917 | 0.8917 |
|  | BGE-Large | 0.8865 | 0.8865 | 0.8865 | 0.8785 | 0.8785 | 0.8785 |

# 4 Discussion

The proposed MMRAG framework is particularly valuable for the biomedical domain, where labeled data is scarce due to privacy constraints and annotation costs. Biomedical NLP tasks require high precision, as errors in entity recognition or relation extraction can lead to misinformation in clinical settings. By integrating retrieval-augmented generation, this framework improves in-context learning by dynamically selecting relevant and diverse examples. The results demonstrate that Top Mode provided the highest F1 scores in precision-sensitive tasks such as NER and RE, demonstrating that high-relevance retrieval improves performance. However, the Diversity Mode achieved more stable generalization across datasets while preserving task relevance. Class Mode, while slightly reducing F1 scores in some cases, ensured broader category representation, which is crucial for classification and multi-relation extraction tasks. Unlike traditional fine-tuning, this approach allows models to adapt to evolving biomedical knowledge. Its flexibility makes it suitable for applications in clinical decision support, drug discovery, and genomic research.

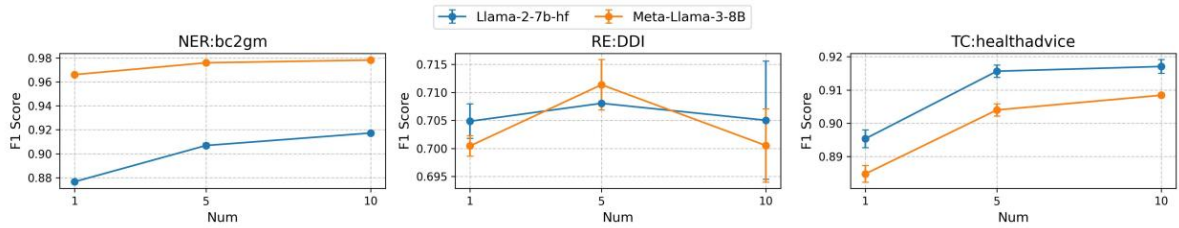

Figure 2. Random Mode (Few-shot) comparison across three datasets.

Figure 2 illustrates the impact of different numbers of examples on model performance in random mode. For the NER (BC2GM) and TC (HealthAdvice) tasks, an increase in the number of examples generally improves performance. However, the performance gain from 5 to 10 examples is noticeably smaller compared to the significant improvement observed from 1 to 5 examples. This suggests that while the jump from 1 to 5 examples provides substantial additional information, the marginal benefit of increasing from 5 to 10 examples is limited. Interestingly, the trend for RE (DDI) is different. The model's performance declines when increasing from 5 to 10 examples. This can be attributed to the random selection of examples, which may include irrelevant or misleading instances. As a result, these less informative or even contradictory examples could lead the model to make incorrect predictions, ultimately degrading performance.

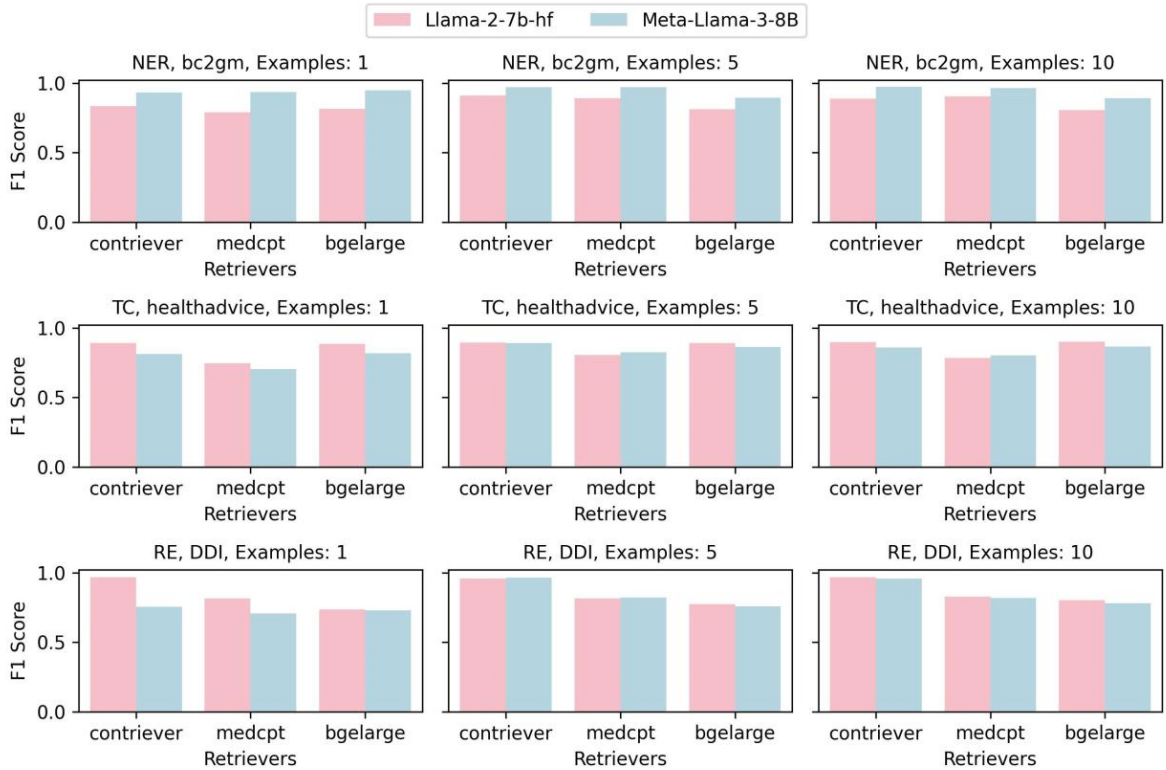

Figure 3. Results of Top mode.

Figure 3 presents the F1-score comparison across different numbers of examples and retrievers in top mode. The results reveal that Llama3 significantly outperforms Llama2 on the NER (BC2GM) task. However, in contrast, Llama2 performs better than Llama3 on the TC (HealthAdvice) and RE (DDI) tasks. In terms of retriever performance, all three retrievers are well-suited for NER (BC2GM). For TC (HealthAdvice), Contriever and BGE-Large outperform MedCPT, indicating their stronger retrieval effectiveness in this task. In the RE (DDI) task, Contriever demonstrates superior performance compared to the other retrievers. Overall, these findings highlight that the effectiveness of different models and retrievers varies across tasks, emphasizing the importance of choosing the appropriate combination for optimal results.

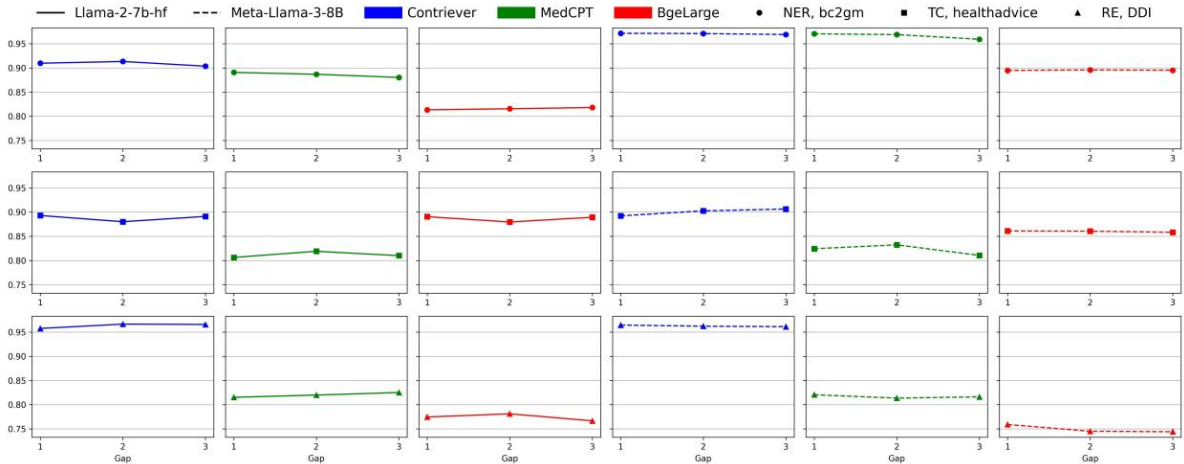

Figure 4. Result of diversity mode with 5 examples

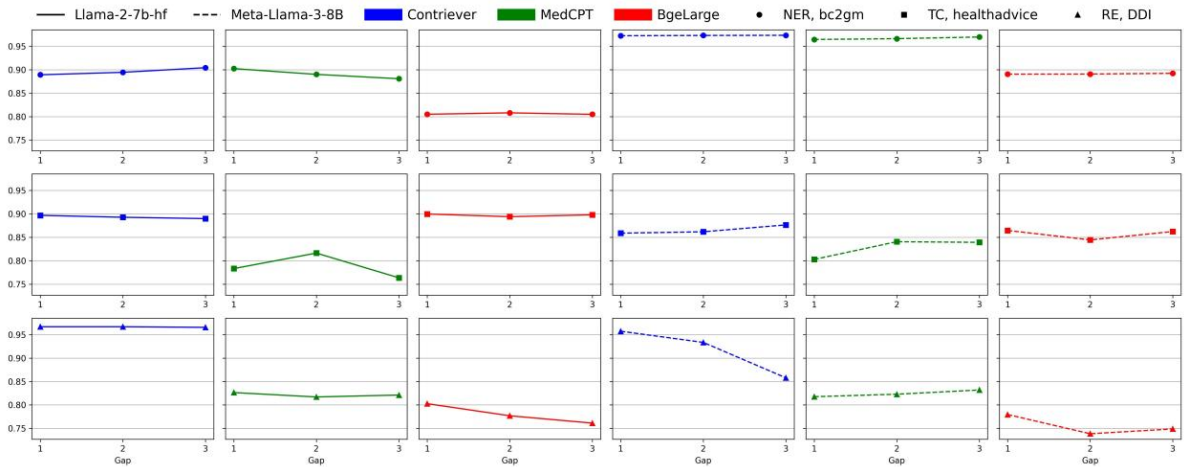

Figure 5. Result of diversity mode with 10 examples

Figures 4 and 5 present the results for Diversity Mode with different gap values using 5 and 10 examples, respectively. Across most experiments, we observe that the gap parameter causes slight fluctuations in performance, but the variations are generally small. This is because Diversity Mode still retrieves examples that are highly similar to the input sentence. However, the nature of these similar examples can influence performance in different ways. If the retrieved examples contain useful and relevant information, they contribute positively, leading to performance improvements. Conversely, if they contain irrelevant or misleading information, they may slightly degrade model performance. Regardless of the gap value, the most similar example is always included as one of the retrieved examples, ensuring that the performance variation remains anchored to the top mode baseline with one example setting.

Class Mode results, as shown in Figure 6, demonstrate its effectiveness in classification and relation extraction tasks. While the absolute F1 scores are slightly lower than those in Top Mode, this approach ensures that the retrieved examples cover a broader range of biomedical categories, providing a more diverse reference set for model inference. The retrieval strategy in Class Mode closely resembles human thinking. In real-world classification tasks, people tend to perform better when they have observed a sufficient number of examples across all categories. However, current LLMs and most NLP algorithms are primarily trained for next-word prediction, which limits their ability to generalize in classification scenarios. Enhancing LLM inference by aligning it more closely with human reasoning processes could significantly improve performance. Therefore, we can anticipate that if future LLMs evolve to better mimic human cognitive patterns, strategies like Class Mode could further enhance their effectiveness in complex classification tasks.

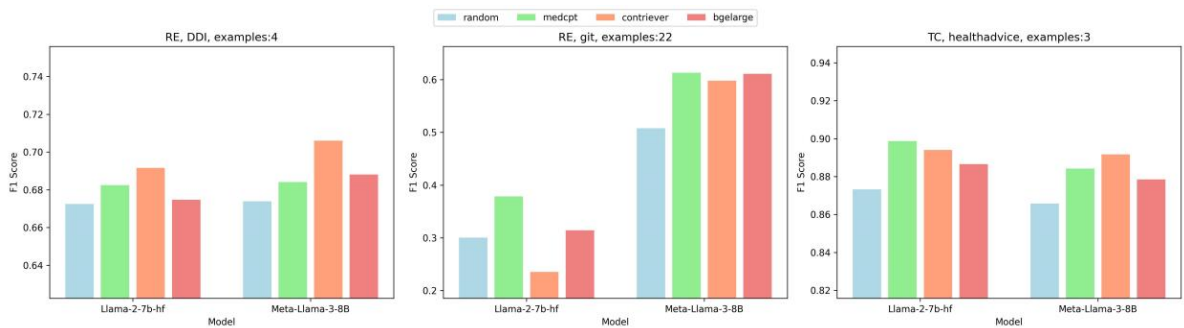

Figure 6. Result of class mode

## 4.1 Error Analysis

Diving deeper into these promising results, our observations of the generated outputs reveal that LLMs mainly suffer from the following two errors:

- Generation Errors: These include misclassifying entities, incorrectly assigning relationships between head and tail entities, and extracting non-entity words as entities. For example, LLMs wrongly identify the word "Genetic" as gene type but it is not.

- Over-Extraction: Due to our strict evaluation criteria, extracting extra entities beyond the ground truth or having boundary errors in entity spans can lead to a decline in performance scores. For example, "CRE-binding-protein" could be tokenized as ['CRE', '-', 'binding', '-', 'protein'] but the LLM recognizes "-" as not part of disease tokens.

## 4.2 Limitations and Future Directions

While this study focuses on example retrieval strategies, several limitations remain:

- Limited model diversity: Although the MMRAG framework is compatible with all generative LLMs, this study only evaluates two LLMs, Llama2 and Llama3. Future work should explore the effectiveness of retrieval strategies across a wider range of models, including newer architectures and domain-specific LLMs.

- Scalability to large datasets: The current framework is applicable to scenarios with large amounts of data. However, to fully leverage big data, future research should focus on developing more efficient retrieval algorithms that optimize performance while maintaining computational feasibility.

By addressing these limitations, retrieval-augmented in-context learning can be further enhanced for broader applications in AI-driven biomedical and other large-scale NLP tasks.

## 5 Conclusion

In this study, we proposed MMRAG, a novel multi-mode retrieval-augmented framework that enhances ICL in biomedical NLP by optimizing example selection. By systematically evaluating four retrieval strategies—Random, Top, Diversity, and Class Mode—our approach demonstrates superior adaptability across multiple biomedical tasks. The results show that retrieval-enhanced selection significantly improves information extraction, particularly in precision-critical tasks. As AI-driven healthcare evolves, MMRAG provides a flexible and efficient method to address data scarcity and privacy challenges, paving the way for future advancements in clinical decision support, drug discovery, and biomedical text processing.

## 6 Acknowledgments

## 7 Author contributions

Zaifu Zhan and Rui Zhang conceptualized and designed the study. Zaifu Zhan curated the data. Zaifu Zhan executed the experiments. Zaifu Zhan drafted the initial manuscript, and all author reviewed and finalized the manuscript. Rui Zhang supervised the whole project.

## 8 Funding Statement

This work was supported by the National Institutes of Health's National Center for Complementary and Integrative Health under grant numbers R01AT009457 and U01AT012871, the National Institute on Aging under grant number R01AG078154, the National Cancer Institute under grant number R01CA287413, the National Institute of Diabetes and Digestive and Kidney Diseases under grant number R01DK115629, and the National Institute on Minority Health and Health Disparities under grant number 1R21MD019134-01. The content is solely the responsibility of the authors and does not represent the official views of the National Institutes of Health.

## 9 Conflicts of interest

The authors state that they have no competing interests to declare.

## FIGURE LEGENDS

Figure 1. An overview of the multi-mode retrieval-augmented generation framework.

Figure 2. Random Mode (Few-shot) comparison across three datasets.

Figure 3. Results of Top mode.

Figure 4. Result of diversity mode with 5 examples

Figure 5. Result of diversity mode with 10 examples

Figure 6. Result of class mode